\documentclass[letterpaper, 10 pt, conference]{ieeeconf}  
\usepackage{cite}
\usepackage{amsmath,amssymb,amsfonts}
\allowdisplaybreaks[4]
\usepackage[caption=false,font=normalsize,labelfont=sf,textfont=sf]{subfig}
\usepackage{algorithmic}
\usepackage{graphicx}
\usepackage{textcomp}
\usepackage{xcolor}
\usepackage{bm}

\IEEEoverridecommandlockouts                              

\title{\LARGE \bf
Collision-Free 6-DoF Trajectory Generation for Omnidirectional Multi-rotor Aerial Vehicle
}

\author{Peiyan Liu, Fengyu Quan, Yueqian Liu, and Haoyao Chen
\thanks{This work was supported in part by the National Natural Science Foundation of China under Grant U1713206 and Grant 61673131. (Corresponding author: Haoyao Chen.)}
\thanks{P.Y. Liu, F.Y. Quan, Y.Q. Liu, and H.Y. Chen* are with the School of Mechanical Engineering and Automation, Harbin Institute of Technology Shenzhen, P.R. China, e-mail:hychen5@hit.edu.cn.}%
}

\begin{document}

\maketitle
\thispagestyle{empty}
\pagestyle{empty}

\begin{abstract}
    As a kind of fully actuated system, omnidirectional multirotor aerial vehicles (OMAVs) has more flexible maneuverability than traditional underactuated multirotor aircraft, and it also has more significant advantages in obstacle avoidance flight in complex environments.
    However, there is almost no way to generate the full degrees of freedom trajectory that can play the OMAVs' potential.
    Due to the high dimensionality of configuration space, it is challenging to make the designed trajectory generation algorithm efficient and scalable.
    This paper aims to achieve obstacle avoidance planning of OMAV in complex environments. A 6-DoF trajectory generation framework for OMAVs was designed for the first time based on the geometrically constrained Minimum Control Effort (MINCO) trajectory generation framework.
    According to the safe regions represented by a series of convex polyhedra, combined with the aircraft's overall shape and dynamic constraints, the framework finally generates a collision-free optimal 6-DoF trajectory.
    The vehicle's attitude is parameterized into a 3D vector by stereographic projection.
    Simulation experiments based on Gazebo and PX4 Autopilot are conducted to verify the performance of the proposed framework.
\end{abstract}

\section{Introduction}
In recent years, with their simple mechanical structure and excellent flight stability, 
multirotor aerial vehicles (MAVs) have stood out from many intelligent robots and come into our life from the laboratory.
However, most traditional MAVs are underactuated systems, which means their translation and rotation dynamics are coupled.
This nature limits the movement ability of traditional MAVs to some extent.
In order to fully explore the potential of MAVs, 
several kinds of omnidirectional MAVs (OMAVs) with decoupled position and attitude control have been developed in recent years.
By changing the rotors' configuration \cite{brescianini2016design} or adding tilting degrees of freedom to the rotors \cite{kamel2018voliro}, 
this kind of MAVs can perform controlled and free rigid body movement, which is impossible for traditional underactuated ones.


Imagine when faced with a narrow straight passage, 
traditional MAVs coupling acceleration with attitude will be most likely unable to pass through it without collision,
while OMAVs can tilt themselves to adapt to the narrow space by controlling the attitude and, simultaneously, control its position to achieve smooth and collision-free crossing.
Such advantages make OMAVs bound to play a great application value in occasions like aerial manipulation and disaster rescue.
In order to better exploit the potential of OMAVs, it is of great significance to design a trajectory generation algorithm framework that can deal with complex environments.
The algorithm needs to take into account the shape and posture of the vehicle to work out extreme scenarios such as narrow passages.
More importantly, we hope this algorithm framework will have excellent computational efficiency, robustness, and extensibility.
To achieve this goal, we are faced with the following challenges:
\begin{itemize}
    \item Due to the high dimensionality of the configuration space $SE(3)$ (6 degrees of freedom), generating trajectories using either search-based or optimization-based methods is prone to high computational time and bad extensibility.
    \item The configuration space $SE(3)$ is a non-Euclidean manifold in which it is difficult to describe the collision-free region. Moreover, it is necessary to find an appropriate way to represent the attitude trajectory to make the trajectory generation problem easy to solve.
\end{itemize}

The existing research on OMAVs mainly focuses on the mechanical structure design and flight control algorithm, but there are few achievements in trajectory planning.
In \cite{brescianini2018computationally}, 6-DoF trajectories for OMAVs are generated efficiently and satisfy certain input constraints using motion primitives.
In \cite{morbidi2018energy}, an energy-efficient trajectory generation method for a tilt-rotor hexarotor UAV is proposed. 
In \cite{pantic2021mesh}, a motion planning method based on Riemannian Motion Policies (RMPs) is proposed. 
This method aims to drive a vehicle to fly to and along a specified surface, 
which is applied to aerial physical interaction.

The above works, without exception, do not take into account the obstacles in the environment.
In \cite{nguyen2016time}, a 6-DoF $SE(3)$ collision-free motion planning method is presented.
RRT is used to search for a collision-free initial path in $SE(3)$ state space, after which time-optimal path parameterization (TOPP) is used to obtain a collision-free 6-DoF trajectory.
However, this method does not show the ability to accommodate arbitrary custom tasks, and it has a pretty long computation time.
It can be seen that the existing works on trajectory planning of OMAVs do not meet our requirements well.

We turn our attention to the field of traditional underactuated multicopters, 
where planning methods in the position space $\mathbb{R}^3$ are relatively mature.
Mellinger et al. \cite{mellinger2011minimumsnap} first smooth the trajectory by minimizing the square integral of the trajectory derivatives,
and a number of efficient schemes have been created based on this idea:
Some are based on the gradient information in the environment \cite{gao2017gradient, zhou2020ego};
Some use intersecting geometry primitives to approximate the free space between the start and goal points, such as \cite{gao2020teach, gao2019flying}, 
and the union of these geometry primitives is called a safe flight corridor (SFC).
However, planning in $\mathbb{R}^3$ only is not enough to exploit the obstacle avoidance potential of OMAVs.
Recently,  an optimization-based trajectory generation framework GCOPTER for multicopters was proposed in \cite{wang2022geometrically}.
By constructing a class of piecewise polynomial trajectories called MINCO and using some constraint processing techniques,
the framework well settles the contradiction between computational efficiency, extensibility, and solution quality.
Based on GCOPTER,  the state-of-the-art whole-body $SE(3)$ trajectory generation framework was proposed in \cite{han2021fast},
In this framework, whole-body safety constraints can be constructed conveniently with 3D free space represented by the convex polyhedron safe flight corridor.

In this paper, inspired by \cite{wang2022geometrically,han2021fast},
an optimization-based collision-free 6-DoF trajectory generation framework considering the vehicle's overall shape and dynamics constraints is designed. 
Just input the SFC, then a collision-free smooth trajectory that satisfies the boundary conditions and dynamic constraints will be obtained efficiently.
To the Best of our Knowledge, our method is the first one that is capable of giving full play to the obstacle avoidance potential of omnidirectional multirotor vehicles.

\section{Preliminaries}

\subsection{Definitions}

\begin{figure}[!t]
    \centering
    \includegraphics[width=3.7in]{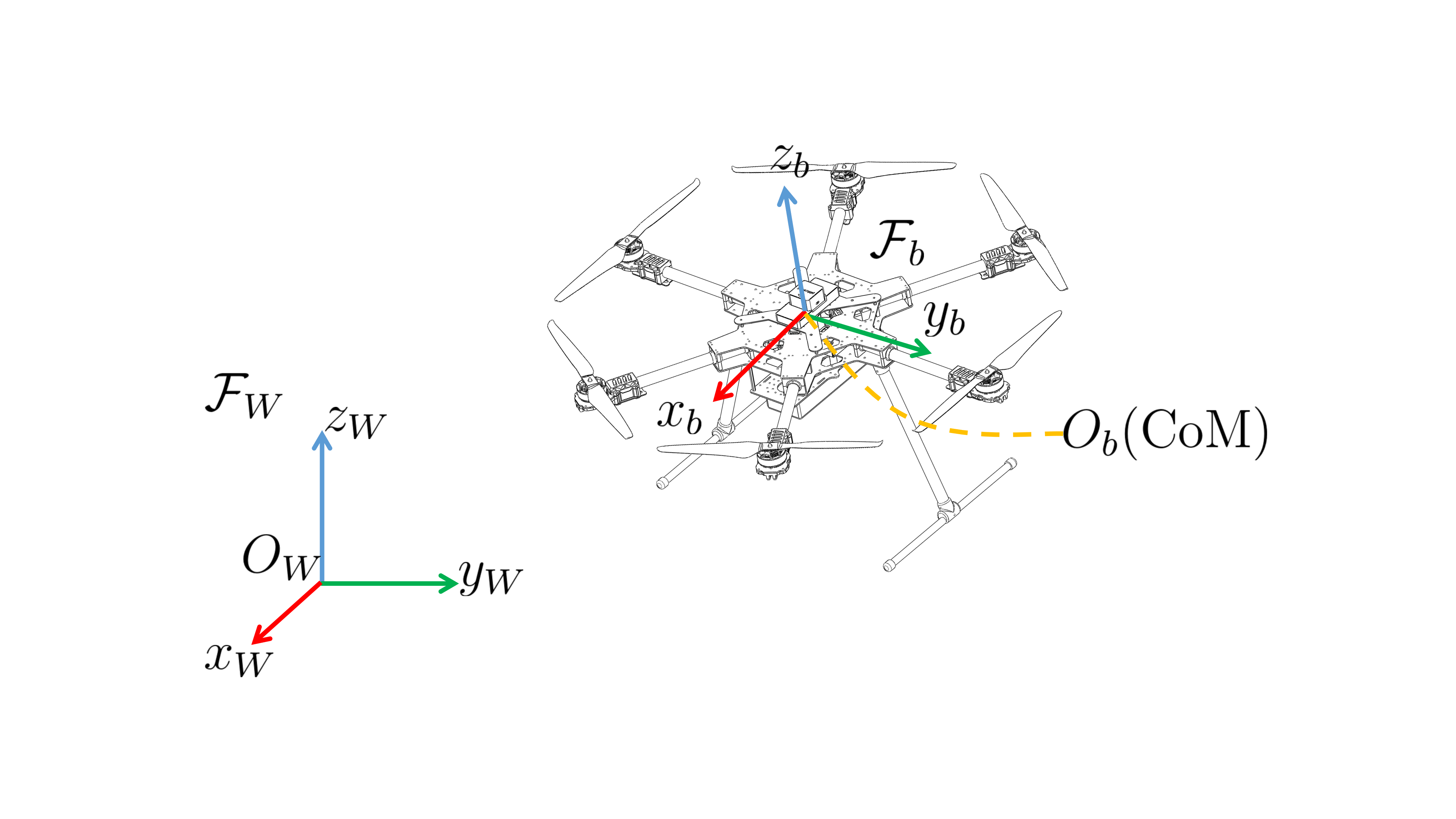}
    \caption{Illustrations of world frame $\mathcal{F}_W$ and body frame $\mathcal{F}_b$.}
\label{fig:frame_definitions}
\end{figure}

The theory of this paper mainly involves two right-handed coordinate systems: the world (inertial) frame $\mathcal{F}_W$ and the body frame $\mathcal{F}_b$ (Fig.~\ref{fig:frame_definitions}).
The world frame uses the ENU coordinate system.
The origin of the body frame coincides with the vehicle's Center of Mass (CoM), 
and the $x_b$ axis points forward, the $y_b$ axis points to the body's left, and the right-hand rule determines the $z_b$ axis.

We denote $\mathbf{a}_A$ the coordinate of a vector $\mathbf{a}$ expressed in frame $\mathcal{F}_A$, and we omit the subscript if $\mathcal{F}_A$ is the world frame $\mathcal{F}_W$.
Denote $\mathbf{R}$ the rotation matrix of the body frame $\mathcal{F}_b$ w.r.t the world frame $\mathcal{F}_W$,
and then $\mathbf{a} = \mathbf{R}\mathbf{a}_b$

\subsection{Differential Flatness}\label{subsec:differential_flatness}
Differential flatness is a fundamental property in the field of robot motion planning, 
which allows us to obtain the system state $\mathbf{x}$ and control input $\mathbf{u}$ from a set of system outputs $\mathbf{z}$ and their finite derivatives, and $\mathbf{z}$ is called flat output.

Differential flatness of underactuated multirotor vehicles have been widely studied.
We will show that OMAVs also have similar properties. 

An OMAV has six independent control degrees of freedom,
we take joint thrust and torque generated by the rotors in body frame $\mathcal{F}_b$ as its control input
\begin{equation}
    \mathbf{u} = 
    \begin{bmatrix}
        \mathbf{f}^{\text{T}}_b & \bm{\tau}^{\text{T}}_b
    \end{bmatrix}^{\text{T}} \in \mathbb{R}^6
    \label{equ:control_input}
\end{equation}
and we select position, attitude, velocity, and angular velocity of body frame $\mathcal{F}_b$ expressed in world frame $\mathcal{F}_W$ as its state variables:
\begin{equation}
    \begin{aligned}
        \mathbf{x} &= 
        \left[
            p_x, p_y, p_z, \phi, \theta, \psi, v_x, v_y, v_z, \omega_x, \omega_y, \omega_z
        \right]^{\text{T}} \\
        &= \begin{bmatrix}
            \mathbf{p}^{\text{T}} & \bm{\varepsilon}^{\text{T}} & \dot{\mathbf{p}}^{\text{T}} & \bm{\omega}^{\text{T}}
        \end{bmatrix}^{\text{T}} \in \mathbb{R}^{12}
    \end{aligned}
    \label{equ:state_variable}
\end{equation}
The system output is given by the position of its CoM and the orientation of $\mathcal{F}_b$ expressed in $\mathcal{F}_W$:
\begin{equation}
    \mathbf{y}(t) = \{\mathbf{p}(t), \mathbf{R}(t)\} \in SE(3)
    \label{equ:trajectory_y}
\end{equation}

It is intuitive to express an attitude trajectory as a function of the rotation matrix w.r.t time $t$.
However, this implies equality constraints which can be troublesome for trajectory generation.
So we consider to parameterize attitude $\mathbf{R}$ as an unconstrained 3D vector.
Denote the attitude trajectory by a curve $\bm{\sigma}(t) : [t_0, t_M] \mapsto \mathbb{R}^3$,
and the corresponding $\mathbf{R}$ at any time is determined by a smooth surjection $\mathbf{R}(\bm{\sigma}) : \mathbb{R}^3 \mapsto SO(3)$.
Then we express the trajectory with full degrees of freedom after attitude parameterization as
\begin{equation}
    \mathbf{z}(t) = 
    \begin{bmatrix}
        \mathbf{p}^{\text{T}}(t) & \bm{\sigma}^{\text{T}}(t)
    \end{bmatrix}^{\text{T}} \in \mathbb{R}^6
    \label{equ:trajectory_z}
\end{equation}

Now according to the relation between the rotation matrix differentiation and the angular velocity, 
the angular velocity of $\mathcal{F}_b$ can be obtained by $\bm{\sigma}$ and its finite derivatives:
\begin{equation}
    \bm{\omega}^{\wedge} = \mathbf{{\omega}}^{\wedge}(\bm{\sigma}, \dot{\bm{\sigma}})
                    = \sum_{i=1}^3 \frac{\partial \mathbf{R}(\bm{\sigma})}{\partial \sigma_i} \dot{\sigma}_i \mathbf{R}(\bm{\sigma})^{\text{T}} 
    \label{equ:angular_velocity_wedge}
\end{equation}
where $(\cdot)^{\wedge} : \mathbb{R}^3 \mapsto \mathfrak{so}(3)$ denotes taking the skew-symmetric matrix of the 3D vector $\cdot$ thus $\mathbf{a} \times \mathbf{b} = \mathbf{a}^{\wedge}\mathbf{b}$. 
Its inverse map is $(\cdot)^{\vee} : \mathfrak{so}(3) \mapsto \mathbb{R}^3$. 
The angular acceleration can be further calculated according to \eqref{equ:angular_velocity_wedge} :
\begin{equation}
    \begin{aligned}
        (\dot{\bm{\omega}})^{\wedge} = \frac{\text{d}\bm{\omega}^{\wedge}}{\text{d}t} &=
        \frac{\text{d}}{\text{d}t}\left(\sum_{i=1}^3 \frac{\partial \mathbf{R}(\bm{\sigma})}{\partial \sigma_i}\dot{\sigma}_i\mathbf{R}^{\text{T}}(\bm{\sigma})\right) \\
        &= \sum_{i=1}^3 \left(\sum_{j=1}^3 \frac{\partial^2 \mathbf{R}}{\partial\sigma_i\partial\sigma_j}\ddot{\sigma}_j\dot{\sigma}_i + \frac{\partial \mathbf{R}}{\partial \sigma_i}\ddot{\sigma}_i\right)\mathbf{R}^{\text{T}} + 
        \dot{\mathbf{R}}\dot{\mathbf{R}}^{\text{T}}
    \end{aligned}
    \label{equ:angular_acceleration_wedge}
\end{equation}
Thus, we obtain the function relation of the state variable $\mathbf{x}$ w.r.t $\mathbf{z}$ and its finite derivatives:
\begin{equation}
    \mathbf{x} = 
    \begin{bmatrix}
        \mathbf{p} \\ \bm{\varepsilon} \\ \dot{\mathbf{p}} \\ \bm{\omega}
    \end{bmatrix} = 
    \Psi_x(\mathbf{z}, \dot{\mathbf{z}}) = 
    \begin{bmatrix}
        \mathbf{p} \\ \bm{\varepsilon}(\bm{\sigma}) \\
        \dot{\mathbf{p}} \\ 
        \left(\sum_{i=1}^3 \frac{\partial \mathbf{R}(\bm{\sigma})}{\partial \sigma_i} \dot{\sigma}_i \mathbf{R}(\bm{\sigma})^{\text{T}}\right)^{\vee}
    \end{bmatrix}
    \label{equ:Psi_x}
\end{equation}
The function relation of the control input $\mathbf{u}$ w.r.t $\mathbf{z}$ and its finite derivatives can be determined by combining \eqref{equ:angular_velocity_wedge} and \eqref{equ:angular_acceleration_wedge} with Newton-Euler equation:
\begin{equation}
    \mathbf{u} = 
    \begin{bmatrix}
        \mathbf{f}_b \\ \bm{\tau}_b
    \end{bmatrix} = 
    \Psi_u(\mathbf{z}, \dot{\mathbf{z}}, \ddot{\mathbf{z}}) = 
    \begin{bmatrix}
        m \mathbf{R}(\bm{\sigma})^{\text{T}} (\ddot{\mathbf{p}} - \mathbf{g}) \\ 
        \mathbf{R}(\bm{\sigma})^{\text{T}} \left(\bm{\omega}^{\wedge}\mathbf{J}(\bm{\sigma})\bm{\omega} + \mathbf{J}(\bm{\sigma})\dot{\bm{\omega}}\right)
    \end{bmatrix}
    \label{equ:Psi_u}
\end{equation}
where $m$ is the mass of the system;
$\mathbf{g} = \begin{bmatrix} 0 & 0 & -9.8\text{m}\cdot\text{s}^{-2}\end{bmatrix}^\text{T}$ is acceleration of gravity in $\mathcal{F}_W$;
$\mathbf{J}_b$ is the system's inertia matrix in $\mathcal{F}_b$ and $\mathbf{J}(\bm{\sigma}) = \mathbf{R}(\bm{\sigma}) \mathbf{J}_b \mathbf{R}(\bm{\sigma})^{\text{T}}$.

\eqref{equ:Psi_x} and \eqref{equ:Psi_u} indicate that our selected trajectory representation $\mathbf{z}(t)$ has properties similar to flat output. 
Therefore, it is convenient to impose constraints on states and inputs in the process of generating 6-DoF $SE(3)$ trajectories for OMAVs.

\section{method}  \label{sec:method}


\subsection{Whole-body Safety Constraint}\label{subsec:safety_constraint}

In our method, we use convex polyhedron SFC $\{\mathcal{P}_i\}_{i=1}^{M_{\mathcal{P}}}$ to describe obstacle-free regions in 3D space connecting the start and goal points. 
Two adjacent convex polyhedra satisfy the connection condition:
\begin{equation}
    \left(\mathcal{P}_{i} \cap \mathcal{P}_{i + 1}\right)^\circ \neq \emptyset, i = 1, \cdots, M_{\mathcal{P}} - 1
    \label{equ:connection_condition_of_sfc}
\end{equation}
where $(\cdot)^\circ$ denotes the interior of the set $\cdot$.

Inspired by \cite{han2021fast}, the shape of the vehicle is approximated by a convex polyhedron $\mathcal{P}_\text{S}$ that wraps the entire vehicle and is fixed to $\mathcal{F}_b$. 
The coordinates of its vertices in $\mathcal{F}_b$ denoted by $\tilde{\bm{v}}_l, l=1, \cdots L_{\text{v}}$ are known constants,.
The safety of the vehicle can be guaranteed as long as the $L_{\text{v}}$ vertices are all in the convex polyhedron $\mathcal{P}$ that represents the safe region, 
We express $\mathcal{P}$ using linear inequalities:
\begin{equation}
    \mathcal{P} = \{\mathbf{p} \in \mathbb{R}^3 \vert \mathbf{n}_k^{\text{T}}\mathbf{p} - d_k \leq 0, \Vert \mathbf{n}_k \Vert_2 = 1, k = 1, \cdots, K\}
    \label{equ:H_representation_of_P}
\end{equation}
which means that the convex polyhedron $\mathcal{P}$ is bounded by $K$ half-spaces.
$\mathbf{n}_k$ is the unit outer normal vector of the $k$-th half-space. 
Then the safety conditions of the vehicle with position $\mathbf{p} \in \mathbb{R}^3$ and attitude $\mathbf{R} \in SO(3)$ can be written as the following inequality constraints
\begin{equation}
    \mathbf{n}_k^{\text{T}}\bm{v}_l - d_k \leq 0, \forall k \in \{1, \cdots, K\}, \forall l \in \{1, \cdots, L_{\text{v}}\}
    \label{equ:safety_condition_at_p_R}
\end{equation}
where $\bm{v}_l = \mathbf{p} + \mathbf{R}\tilde{\bm{v}}_l, l=1, \cdots L_{\text{v}}$ are coordinates of vertices of $\mathcal{P}_\text{S}$ in $\mathcal{F}_W$:

There are many ways to generate SFC at present, 
such as RILS \cite{liu22017planning} used in the implementation of this paper. 
After the SFC is generated, 
it will be treated as a set of linear inequality constraints during trajectory optimization.

\subsection{6-DoF Trajectory Optimization} \label{subsec:problem_formulation}
In order to obtain and control the corresponding point and the derivatives at any time on the trajectory $\mathbf{z}(t)$, 
we express $\mathbf{z}(t)$ as piecewise polynomials:
\begin{equation}
    \mathbf{z}(t) = \mathbf{c}_i^{\text{T}}\bm{\beta}(t - t_{i - 1}), \forall t \in [t_{i - 1}, t_i], i = 1, \cdots, M
    \label{equ:kth_M_piece_polynomial}
\end{equation}
where $\mathbf{c}_i \in \mathbb{R}^{(k+1) \times 6}$ is the coefficient matrix of the $i$-th piece and
$\bm{\beta}(\alpha) = \begin{bmatrix}1 & \alpha & \cdots & \alpha^k\end{bmatrix}^{\text{T}} \in \mathbb{R}^{(k+1)}$.
The coefficient matrix of the whole trajectory is $\mathbf{c} = \begin{bmatrix}\mathbf{c}_1^{\text{T}} & \cdots & \mathbf{c}_M^{\text{T}}\end{bmatrix}^{\text{T}} \in \mathbb{R}^{M(k+1) \times 6}$.
The time for each segment is $\mathbf{T} = \begin{bmatrix}t_1 - t_0 & \cdots & t_M - t_{M-1}\end{bmatrix}^{\text{T}} = \begin{bmatrix}T_1 & \cdots & T_M\end{bmatrix}^{\text{T}} \in \mathbb{R}_+^M$.

Trajectory optimization is to find the appropriate $\mathbf{c}$ and $\mathbf{T}$ that minimize a given objective function,
but the high dimensionality of $\mathbf{c}$ may bring difficulties to it.
We use MINCO trajectory representation to reduce the dimension of parameters of the piecewise polynomial $\mathbf{z}(t)$ to the intermediate points $\mathbf{q} = \begin{bmatrix}\mathbf{q}_1 & \cdots & \mathbf{q}_{M-1}\end{bmatrix} \in \mathbb{R}^{6 \times (M - 1)}$ and $\mathbf{T}$,
where $\mathbf{q}_i = \mathbf{z}(t_i),i = 1, \cdots, M-1$.
After specifying $\mathbf{q}$ $\mathbf{T}$, the start condition $\bar{\mathbf{z}}_o = \begin{bmatrix} \mathbf{z}(t_0)^\text{T} & \cdots & \mathbf{z}^{(s-1)}(t_0)^\text{T}
\end{bmatrix}^\text{T} \in \mathbb{R}^{6s}$, and the end condition $\bar{\mathbf{z}} _f = \begin{bmatrix} \mathbf{z}(t_M)^\text{T} & \cdots & \mathbf{z}^{(s-1)}(t_M)^\text{T}
\end{bmatrix}^\text{T} \in \mathbb{R}^{6s}$, the coefficient $\mathbf{c}$ will be calculated in an efficient way according to Theorem 2 in \cite{wang2022geometrically}, 
and the degree $k = 2s - 1, s \in \mathbb{N}_+$.

For the convenience of the following statement, 
we divide the coefficient matrix $\mathbf{c}$ and the intermediate points $\mathbf{q}$ into position blocks and attitude blocks:
$ \mathbf{c}_i = 
\begin{bmatrix}
    \mathbf{c}_i^{p} & \mathbf{c}_i^{\sigma}
\end{bmatrix}, \mathbf{c}_i^{p}, \mathbf{c}_i^{\sigma} \in \mathbb{R}^{2s \times 3} $; 
$ \mathbf{c} = 
\begin{bmatrix}
    \mathbf{c}^{p} & \mathbf{c}^{\sigma}
\end{bmatrix}, \mathbf{c}^{p}, \mathbf{c}^{\sigma} \in \mathbb{R}^{2Ms \times 3} $; 
$ \mathbf{q}_i = 
\begin{bmatrix}
    {\mathbf{q}_i^{p}}^{\text{T}} & {\mathbf{q}_i^{\sigma}}^{\text{T}}
\end{bmatrix}^{\text{T}}, \mathbf{q}_i^{p}, \mathbf{q}_i^{\sigma} \in \mathbb{R}^{3} $; 
$ \mathbf{q} = 
\begin{bmatrix}
    {\mathbf{q}^{p}}^{\text{T}} & {\mathbf{q}^{\sigma}}^{\text{T}}
\end{bmatrix}^{\text{T}}, \mathbf{q}^{p}, \mathbf{q}^{\sigma} \in \mathbb{R}^{3 \times (M-1)} $.

Our framework aims to obtain a smooth trajectory $\mathbf{z}^*(t): [t_0, t_M] \mapsto \mathbb{R}^6$ that at least satisfies the boundary conditions, the dynamic constraints, and the safety constraints for any $t\in[t_0, t_M]$.
The smoothness of a trajectory is measured using the integral of $\Vert \mathbf{z}^{(s)}(t) \Vert_2^2$ w.r.t time, 
then specifically, the original form of our trajectory optimization problem is as follows:
\begin{subequations}
    \begin{align}
        \min_{\mathbf{q}, \mathbf{T}} &\int_{t_0}^{t_M} \Vert \mathbf{z}^{(s)}(t) \Vert_2^2 \text{d}t + k_\rho \Vert \mathbf{T} \Vert_1 \label{equ:origin_problem_obj}\\
        s.t. \quad 
            & \begin{aligned}
                & \mathbf{z}(t) = \mathbf{c}_i^{\text{T}}(\mathbf{q}, \mathbf{T})\bm{\beta}(t - t_{i - 1}), \forall t \in [t_{i - 1}, t_i], \\
                & i = 1, \cdots, M \label{equ:origin_problem_cons_1}
            \end{aligned} \\
             & \mathbf{z}^{[s - 1]}(t_0) = \bar{\mathbf{z}}_o, \mathbf{z}^{[s - 1]}(t_M) = \bar{\mathbf{z}}_f \label{equ:origin_problem_cons_2}\\
             &\begin{aligned}
                & \mathbf{q}_{k_i}^p \in \left(\mathcal{P}_{i} \cap \mathcal{P}_{i + 1}\right)^\circ, 1 \leq k_i < k_{i + 1} \leq M - 1, \\
                & i = 1, \cdots, M_{\mathcal{P}} - 1 \label{equ:origin_problem_cons_3}
             \end{aligned} \\
             & \begin{aligned}
                & \mathbf{q}_j^p \in \mathcal{P}_1, \text{if} \ 1 \leq j < k_1;  \\
                & \mathbf{q}_j^p \in \mathcal{P}_{M_\mathcal{P}}, \text{if} \ k_{M_{\mathcal{P}} - 1} < j \leq M - 1 \label{equ:origin_problem_cons_4}
             \end{aligned} \\
             & \mathbf{q}_j^p \in \mathcal{P}_i, \text{if} \ k_{i-1} < j < k_i, i = 2, \cdots, M_{\mathcal{P}} - 1 \label{equ:origin_problem_cons_5}\\
             & \mathbf{T} \succ \mathbf{0} \label{equ:origin_problem_cons_6}\\
             & \Vert \dot{\mathbf{p}}(t) \Vert_2 \leq v_{\text{max}}, \forall t \in [t_0, t_M] \label{equ:origin_problem_cons_7}\\
             & \Vert \ddot{\mathbf{p}}(t) \Vert_2 \leq a_{\text{max}}, \forall t \in [t_0, t_M] \label{equ:origin_problem_cons_8}\\
             & \Vert \bm{\omega}(t) \Vert_2 \leq \omega_{\text{max}}, \forall t \in [t_0, t_M] \label{equ:origin_problem_cons_9}\\
             & \bm{v}_l(t) \in \mathcal{P}^i, \forall t \in [t_{i - 1}, t_i], i = 1, \cdots M; l = 1, \cdots, L_{\text{v}} \label{equ:origin_problem_cons_10}
    \end{align}
\end{subequations}

The first term of the objective function \eqref{equ:origin_problem_obj} is the smoothness term, denoted as $J$;
the second term is the time regularization term. 
\eqref{equ:origin_problem_cons_3} to \eqref{equ:origin_problem_cons_5} binds the position intermediate points to a specific region in the SFC, 
this constraint can prevent the trajectory from deviating from the SFC during the optimization process. 
\eqref{equ:origin_problem_cons_6} ensures that the time allocated to each piece is not zero.
\eqref{equ:origin_problem_cons_7} to \eqref{equ:origin_problem_cons_10} are continuous-time constraints.
Specifically, \eqref{equ:origin_problem_cons_7} to \eqref{equ:origin_problem_cons_9} are dynamic constraints;
\eqref{equ:origin_problem_cons_10} represents the security constraints,
where we assign each piece to a convex polyhedron member of SFC.
$\mathcal{P}^i$ denotes the convex polyhedron to which the $i$-th piece is assigned, and we represent it as:
\begin{equation}
    \mathcal{P}^i = \{\mathbf{p} \in \mathbb{R}^3 \vert \mathbf{n}_{i, k}^{\text{T}}\mathbf{p} - d_{i, k} \leq 0, \Vert \mathbf{n}_{i, k} \Vert_2 = 1, k = 1, \cdots, K_i\}
\end{equation}

For the selection of order $s$, according to \eqref{equ:Psi_u},
$\mathbf{z}(t)$ needs to be at least three-order continuously differentiable to ensure the smoothness of the control input. 
That is, $2s - 1 \geq 4$, which gives $s \geq 3$.

To deal with the constraints which may bring trouble to the optimization process, 
as shown in \cite{wang2022geometrically}, we can use some techniques to handle them.
The continuous-time constraints \eqref{equ:origin_problem_cons_7} to \eqref{equ:origin_problem_cons_10} can be softened as integral penalty terms and incorporated into the objective function; 
this method allows users to add any custom constraints for specific tasks rather than just those listed, 
which significantly enhances the extensibility of the framework.
The spatial constraints \eqref{equ:origin_problem_cons_3} to \eqref{equ:origin_problem_cons_5} and time constraint \eqref{equ:origin_problem_cons_6} can be eliminated by using appropriate diffeomorphism $\mathbf {q}(\bm{\xi})$ and $\mathbf{T} (\bm{\tau})$.
Finally, what we need to solve is an unconstrained optimization problem as follows:
\begin{equation}
    \begin{aligned}
        \min_{\bm{\xi}, \mathbf{q}^{\sigma}, \bm{\tau}}
        &J(\mathbf{q}^p(\bm{\xi}), \mathbf{q}^{\sigma}, \mathbf{T}(\bm{\tau})) + k_\rho \Vert \mathbf{T}(\bm{\tau}) \Vert_1 \\
        &+ \mathcal{W}_v \sum_{i=1}^M \sum_{j=1}^{\kappa} \mathcal{V}\left(\left\| \dot{\mathbf{p}}\left(\hat{t}_{ij}\right) \right\|_2^2 - v_{\text{max}}^2\right) \frac{T_i}{\kappa} \\
        &+ \mathcal{W}_a\sum_{i=1}^M \sum_{j=1}^{\kappa} \mathcal{V}\left(\left\| \ddot{\mathbf{p}}\left(\hat{t}_{ij}\right) \right\|_2^2 - a_{\text{max}}^2\right) \frac{T_i}{\kappa} \\
        &+ \mathcal{W}_{\omega} \sum_{i=1}^M \sum_{j=1}^{\kappa} \mathcal{V}\left(\left\| \bm{\omega}\left(\hat{t}_{ij}\right) \right\|_2^2 - \omega_{\text{max}}^2\right) \frac{T_i}{\kappa} \\
        &+ \mathcal{W}_c \sum_{i=1}^M \sum_{j=1}^{\kappa} \sum_{l=1}^{L_{\text{v}}} \sum_{k=1}^{K_i} \mathcal{V}\left(\mathbf{n}_{i, k}^{\text{T}}\bm{v}_l\left(\hat{t}_{ij}\right) - d_{i, k}\right) \frac{T_i}{\kappa}
    \end{aligned}
    \label{equ:final_uncons_opt_problem}
\end{equation}
where $\mathcal{W}_\star$ is the weight of the corresponding penalty term;
$\mathcal{V}(\cdot)=\max(\cdot, 0)^3$ measures the degree of constraint violation on the trajectory at the sampling time $\hat{t}_{ij} = t_{i-1} + \frac{j}{\kappa}T_i$;
$\kappa \in \mathbb{N}_+$ controls the resolution of numerical integration.

The unconstrained optimization problem \eqref{equ:final_uncons_opt_problem} can be solved by quasi-Newton methods.
It should be noted that the continuous time constraints softened and discretized above may be slightly violated, 
but in most cases, these violations are within the acceptable range.

\subsection{Gradient Calculation} \label{subsec:general_gradient_calculation}
To solve the unconstrained optimization problem \eqref{equ:final_uncons_opt_problem} efficiently, we need to obtain the gradient of the objective function w.r.t the object variables. 
We first find the gradient w.r.t $\mathbf{c}$ and $\mathbf{T}$, 
then the gradient w.r.t object variables $\bm{\xi}$, $\mathbf{q}^\sigma$, and $\bm{\tau}$ can be obtained using the method proposed in \cite{wang2022geometrically}.

Gradient of the time-regularized smoothness term $\mathcal{J} = J + k_\rho\Vert\mathbf{T}\Vert_1$ can be calculated as follows:
\begin{align}
    & \frac{\partial \mathcal{J}}{\partial \mathbf{c}_i} = 2 \left(\int_0^{T_i} \bm{\beta}^{(s)}(t)\bm{\beta}^{(s)}(t)^{\text{T}} \text{d}t\right) \mathbf{c}_i \\
    & \frac{\partial \mathcal{J}}{\partial T_i} = \mathbf{c}_i^{\text{T}}\bm{\beta}^{(s)}(t)\bm{\beta}^{(s)}(t)^{\text{T}}\mathbf{c}_i + k_\rho
\end{align}

For the penalty terms, each of them is the summary of several sub-penalty terms. 
When the constraint corresponding to a sub-penalty term is satisfied, the term and its gradient are always 0. 
So we only need to consider the sub-penalty terms that violate the corresponding constraints,
and we denote them as 
\begin{align}
    & P_{v_{ij}} = \left(\Vert \dot{\mathbf{p}}(\hat{t}_{ij}) \Vert_2^2 - v_{\text{max}}^2\right)^3 \frac{T_i}{\kappa} = \mathcal{G}_{v_{ij}}^3 \frac{T_i}{\kappa} \\
    & P_{a_{ij}} = \left(\Vert \ddot{\mathbf{p}}(\hat{t}_{ij}) \Vert_2^2 - a_{\text{max}}^2\right)^3 \frac{T_i}{\kappa} = \mathcal{G}_{a_{ij}}^3 \frac{T_i}{\kappa} \\
    & P_{\omega_{ij}} = \left(\Vert \bm{\omega}(\hat{t}_{ij}) \Vert_2^2 - \omega_{\text{max}}^2\right)^3 \frac{T_i}{\kappa} = \mathcal{G}_{\omega_{ij}}^3 \frac{T_i}{\kappa} \\
    & P_{c_{ijlk}} = \left(\mathbf{n}_{i, k}^{\text{T}} \left(\mathbf{p}(\hat{t}_{ij}) + \mathbf{R}(\hat{t}_{ij})\tilde{\bm{v}}_l\right) - d_{i, k}\right)^3 \frac{T_i}{\kappa} = \mathcal{G}_{c_{ijlk}}^3 \frac{T_i}{\kappa}
\end{align}

Obviously, the penalty terms corresponding to the $i$-th piece are only related to $\mathbf{c}_i$ and $T_i$,
so we only need to calculate their gradient w.r.t $\mathbf{c}_i$ and $T_i$.
The analytical expression is given as follows:
\begin{subequations}
    \allowdisplaybreaks
    \begin{align}
        & \frac{\partial P_{v_{ij}}}{\partial \mathbf{c}_i} = \begin{bmatrix}
            \frac{6T_i}{\kappa} \mathcal{G}_{v_{ij}}^2 \bm{\beta}^{(1)}(\frac{j}{\kappa}T) \dot{\mathbf{p}}^{\text{T}} &
            \mathbf{0}_{2s \times 3}
        \end{bmatrix} \\
        & \frac{\partial P_{v_{ij}}}{\partial T_i} = \frac{\mathcal{G}_{v_{ij}}^2}{\kappa} \left(\mathcal{G}_{v_{ij}} + \frac{6jT_i}{\kappa}\dot{\mathbf{p}}^{\text{T}}\ddot{\mathbf{p}}\right) \\
        & \frac{\partial P_{a_{ij}}}{\partial \mathbf{c}_i} = \begin{bmatrix}
            \frac{6T_i}{\kappa} \mathcal{G}_{a_{ij}}^2 \bm{\beta}^{(2)}(\frac{j}{\kappa}T) \ddot{\mathbf{p}}^{\text{T}} &
            \mathbf{0}_{2s \times 3}
        \end{bmatrix} \\
        & \frac{\partial P_{a_{ij}}}{\partial T_i} = \frac{\mathcal{G}_{a_{ij}}^2}{\kappa} \left(\mathcal{G}_{a_{ij}} + \frac{6jT_i}{\kappa}\ddot{\mathbf{p}}^{\text{T}}\dddot{\mathbf{p}}\right) \\
        & \frac{\partial P_{\omega_{ij}}}{\partial \mathbf{c}_i} = \begin{bmatrix}
            \mathbf{0}_{2s \times 3} &
            \frac{6T_i}{\kappa} \mathcal{G}_{\omega_{ij}}^2 \left(\omega_x \frac{\partial \omega_x}{\partial \mathbf{c}_i^\sigma} + \omega_y \frac{\partial \omega_y}{\partial \mathbf{c}_i^\sigma} + \omega_z \frac{\partial \omega_z}{\partial \mathbf{c}_i^\sigma}\right) \label{equ:Pomg_d_cis}
        \end{bmatrix} \\
        & \frac{\partial P_{\omega_{ij}}}{\partial T_i} = \frac{\mathcal{G}_{\omega_{ij}}^2}{\kappa} \left(\mathcal{G}_{\omega_{ij}} + \frac{6T_i}{\kappa}\bm{\omega}^{\text{T}}\frac{\partial \bm{\omega}}{\partial T_i}\right) \label{equ:Pomg_d_Ti} \\
        & \frac{\partial P_{c_{ijlk}}}{\partial \mathbf{c}_i^p} = \frac{3T_i}{\kappa} \mathcal{G}_{c_{ijlk}}^2 \bm{\beta}(\frac{j}{\kappa}T_i)\mathbf{n}_{i, k}^{\text{T}} \\
        & \left[\frac{\partial P_{c_{ijlk}}}{\partial \mathbf{c}_i^\sigma}\right]_{m, n} = \frac{3T_i}{\kappa} \mathcal{G}_{c_{ijlk}}^2 \text{tr}\{\tilde{\bm{v}}_l\mathbf{n}_{i, k}^{\text{T}}\frac{\partial \mathbf{R}}{\partial [\mathbf{c}_i^\sigma]_{m,n}}\} \label{equ:Pc_d_cis}\\
        & \begin{aligned}
            & \frac{\partial P_{c_{ijlk}}}{\partial T_i} = \\
            & \frac{\mathcal{G}_{c_{ijlk}}^2}{\kappa} \left[\mathcal{G}_{c_{ijlk}} + 3T_i\left(\frac{j}{\kappa}\mathbf{n}_{i, k}^{\text{T}}{\mathbf{c}_i^p}^{\text{T}}\bm{\beta}^{(1)}(\frac{j}{\kappa}T_i) + \text{tr}\{\tilde{\bm{v}}_l\mathbf{n}_{i, k}^{\text{T}}\frac{\partial \mathbf{R}}{\partial T_i}\}\right)\right] \label{equ:Pc_d_Ti}
        \end{aligned}
    \end{align}
\end{subequations}
where $\mathbf{p}$, $\mathbf{R}$, $\bm{\omega}$, and their derivatives are taken as the values at $\hat{t}_{ij}$;
$[\cdot]_{m, n}$ is the element of matrix $\cdot$ with row index $m$ and column index $n$.

Since the attitude-related quantities such as $\mathbf{R}$ and $\bm{\omega}$ are closely related to $\bm{\sigma}$, 
the process of evaluating these penalty terms and their gradient will differ depending on the attitude parameterization method.
Various reasonable attitude parameterization methods can be selected to generate the attitude trajectory, as long as $\mathbf{R}(\bm{\sigma})$ is a smooth surjective.

\subsection{Attitude Parameterization}
As can be seen from \ref{subsec:problem_formulation} and \ref{subsec:general_gradient_calculation},
the difference brought by different attitude parameterization methods to problem \eqref{equ:final_uncons_opt_problem} is only reflected in the penalty terms.
Choosing an appropriate attitude parameterized map $\mathbf{R}(\bm{\sigma})$ is very important for trajectory generation.

There are several commonly used ways to parameterize an attitude in $SO(3)$ as a vector in $\mathbb{R}^3$,
such as Euler angles and Lie algebras.
However, since $\bm{\sigma}$ is a free vector in $\mathbb{R}^3$ and we do not impose any hard constraints on it, 
it will be ambiguous if $\bm{\sigma}$ represents Euler angles or Lie algebras. 
Two very different $\bm{\sigma}$ values will most likely correspond to the same attitude.
Moreover, Euler Angle representation has the problem of gimbal lock. 

Considering these shortcomings, 
parameterizing attitude as Euler angles or Lie algebras lacks rationality in scenarios where polynomial interpolation is used.
In our implementation, 
we choose a parameterization method based on quaternion and stereographic projection, which is more rational as shown in \cite{terzakis2014quaternion}.
This method uses the homeomorphism between the hyperplane $\mathbb{R}^3$ and the hypersphere $\mathbb{S}^3$ with one pole removed, 
A stereographic projection maps an arbitrary vector $\bm{\sigma} \in \mathbb{R}^3$ as a unit quaternion $\mathbf{Q} = \begin{bmatrix}w & x & y & z\end{bmatrix}^\text{T} = \begin{bmatrix}w & \mathbf{r}^{\text{T}}\end{bmatrix}^\text{T}$.
If the pole is chosen as $\mathbf{Q}_{N} = \begin{bmatrix}1 & 0 & 0 & 0\end{bmatrix}^\text{T}$,
the map is as follow:
\begin{equation}
    \mathbf{Q}(\bm{\sigma}) = 
    \begin{bmatrix}
        \frac{\bm{\sigma}^{\text{T}}\bm{\sigma} - 1}{\bm{\sigma}^{\text{T}}\bm{\sigma} + 1} &
        \frac{2\bm{\sigma}^{\text{T}}}{\bm{\sigma}^{\text{T}}\bm{\sigma} + 1}
    \end{bmatrix}^{\text{T}} \in \mathbb{S}^3 \backslash \{\mathbf{Q}_N\}, \forall \bm{\sigma} \in \mathbb{R}^3
    \label{equ:stereographic_projection}
\end{equation}
We can see that the map $\mathbf{Q}(\bm{\sigma}):\mathbb{R}^3 \mapsto \mathbb{S}^3 \backslash \{\mathbf{Q}_N\}$ is smooth and one-to-one.
Each attitude has at most two distinct $\bm{\sigma}$ counterparts ($\mathbf{R} = \mathbf{I}$ corresponds only to the origin of $\mathbb{R}^3$), 
which greatly reduces the possibility of ambiguity.

Now we can calculate the final attitude $\mathbf{R}(\bm{\sigma})$ using the relation between rotation matrices and unit quaternions.
The angular velocity expressed in $\mathcal{F}_W$ can be calculated as:
\begin{equation}
    \bm{\omega} = 2\mathbf{U}\dot{\mathbf{Q}} = 2\mathbf{U}\mathbf{G}^{\text{T}}\dot{\bm{\sigma}}
    \label{equ:omega}
\end{equation}
where $\mathbf{U} = \begin{bmatrix} -\mathbf{r} & w\mathbf{I} + \mathbf{r}^{\wedge}\end{bmatrix} \in \mathbb{R}^{3 \times 4}$
 and $\mathbf{G} = 
 \begin{bmatrix}
     \frac{\partial w}{\partial \bm{\sigma}} & 
     \frac{\partial x}{\partial \bm{\sigma}} & 
     \frac{\partial y}{\partial \bm{\sigma}} & 
     \frac{\partial z}{\partial \bm{\sigma}}
 \end{bmatrix} \in \mathbb{R}^{3 \times 4}$.
Then we can calculate the attitude-related penalty terms.

As for gradient calculation, 
Just calculate the derivatives of $\mathbf{R}$ and $\bm{\omega}$ w.r.t $\bm{\mathbf{c}}_i^{\bm{\sigma}}$ and $\mathbf{T}$, 
then substitute it into \eqref{equ:Pomg_d_cis}, \eqref{equ:Pomg_d_Ti}, \eqref{equ:Pc_d_cis} and \eqref{equ:Pc_d_Ti}.
At the relative sampling time $\bar{t} = jT_i/\kappa$, 
the gradient of the quaternion $\mathbf{Q}$ w.r.t the attitude coefficient matrix $\mathbf{c}_i^{\sigma}$ is
\begin{equation}
    \allowdisplaybreaks
    \begin{aligned}
        \frac{\partial \alpha}{\partial \mathbf{c}_i^\sigma} &= 
        \sum_{k=1}^3 \frac{\partial \alpha}{\partial \sigma_k}\frac{\partial \sigma_k}{\partial \mathbf{c}_i^\sigma} =
        \sum_{k=1}^3 \frac{\partial \alpha}{\partial \sigma_k}\bm{\beta}(\bar{t})\mathbf{e}_k^\text{T} \\ &  = 
        \begin{bmatrix}
            \frac{\partial \alpha}{\partial \sigma_1}\bm{\beta}(\bar{t}) &
            \frac{\partial \alpha}{\partial \sigma_2}\bm{\beta}(\bar{t}) &
            \frac{\partial \alpha}{\partial \sigma_3}\bm{\beta}(\bar{t})
        \end{bmatrix}
    \end{aligned}
    \label{equ:Q_d_cis}
\end{equation}
where $\alpha = w, x, y, z$;
$\mathbf{e}_k$ denotes the $k$-th column of a 3x3 identity matrix.
According to \eqref{equ:omega}, we further let
\begin{equation}
    \bm{\Gamma} = 
    \begin{bmatrix}
        \bm{\gamma}_x & \bm{\gamma}_y & \bm{\gamma}_z
    \end{bmatrix}^\text{T} = 
    \mathbf{UG}^\text{T}
    \label{equ:Gamma}
\end{equation}
and we can derive that 
\newcommand{\pdiff}[2]{\frac{\partial #1}{\partial #2}}
\begin{align}
    &\frac{\partial \omega_{x}}{\partial \mathbf{c}_i^\sigma} = 2\sum_{k=1}^3 \left(\bm{\beta}^{(1)}(\bar{t})\mathbf{e}_k^\text{T}\gamma_{xk} + \dot{\sigma}_k\frac{\partial \gamma_{xk}}{\partial \mathbf{c}_i^\sigma}\right) \\
    &\frac{\partial \omega_{y}}{\partial \mathbf{c}_i^\sigma} = 2\sum_{k=1}^3 \left(\bm{\beta}^{(1)}(\bar{t})\mathbf{e}_k^\text{T}\gamma_{yk} + \dot{\sigma}_k\frac{\partial \gamma_{yk}}{\partial \mathbf{c}_i^\sigma}\right) \\
    &\frac{\partial \omega_{z}}{\partial \mathbf{c}_i^\sigma} = 2\sum_{k=1}^3 \left(\bm{\beta}^{(1)}(\bar{t})\mathbf{e}_k^\text{T}\gamma_{zk} + \dot{\sigma}_k\frac{\partial \gamma_{zk}}{\partial \mathbf{c}_i^\sigma}\right)
\end{align}
To find the derivative of the angular velocity w.r.t $T_i$, 
we first find the derivative of $\mathbf{Q}$ w.r.t $T_i$ along with that of $\mathbf{U}$:
\begin{equation}
    \pdiff{\mathbf{Q}}{T_i} = \frac{j}{\kappa} \mathbf{G}^\text{T} \dot{\bm{\sigma}}
    \label{equ:Q_d_Ti}
\end{equation}
We denote the Hessian of $\mathbf{Q}$ w.r.t $\bm{\sigma}$ as $\mathbf{H}_\alpha,\alpha = w, x, y, z$,
then the derivative of $\mathbf{G}$ w.r.t $T_i$ is
\begin{equation}
    \pdiff{\mathbf{G}}{T_i} = 
    \frac{j}{\kappa} 
    \begin{bmatrix}
        \mathbf{H}_w^\text{T}\dot{\bm{\sigma}} &
        \mathbf{H}_x^\text{T}\dot{\bm{\sigma}} & 
        \mathbf{H}_y^\text{T}\dot{\bm{\sigma}} & 
        \mathbf{H}_z^\text{T}\dot{\bm{\sigma}} 
    \end{bmatrix}
\end{equation}
Then 
\begin{equation}
    \pdiff{\dot{\mathbf{Q}}}{T_i} = 
    \left(\pdiff{\mathbf{G}}{T_i}\right)^\text{T}\dot{\bm{\sigma}} + 
    \frac{j}{\kappa}\mathbf{G}^\text{T}\ddot{\bm{\sigma}}
\end{equation}
The derivative of $\bm{\omega}_b$ w.r.t $T_i$ can be obtained in conjunction with \eqref{equ:omega} as follows:
\begin{equation}
    \pdiff{\bm{\omega}}{T_i} = 
    \frac{\partial}{\partial T_i}(2\mathbf{U}\dot{\mathbf{Q}}) = 
    2\left(\pdiff{\mathbf{U}}{T_i}\dot{\mathbf{Q}} + \mathbf{U}\pdiff{\dot{\mathbf{Q}}}{T_i}\right)
\end{equation}
The derivatives of the rotation matrix $\mathbf{R}$ w.r.t $\mathbf{c}_i^\sigma$ and $T_i$ can be obtained according to $\mathbf{R}(\mathbf{Q})$, \eqref{equ:Q_d_cis} and \eqref{equ:Q_d_Ti}.

\section{Results}
\begin{figure}[!t]
    \centering
    \includegraphics[width=2.5in]{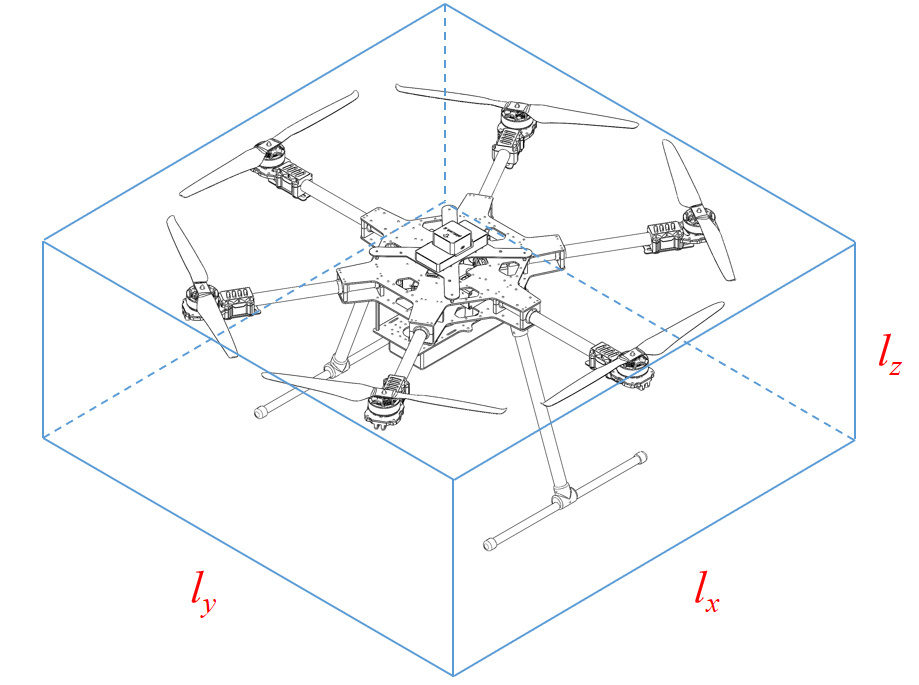}
    \caption{The cuboid used to approximate the shape of vehicle, of which the three symmetry axes are axes of $\mathcal{F}_b$.}
    \label{fig:approximate_cuboid}
\label{fig1}
\end{figure}
This section takes a tilt-rotor omnidirectional hexarotor vehicle (hereinafter referred to as OmniHex) as the research object. 
Firstly, the trajectory generation experiment is carried out in virtual environments with randomly distributed obstacles to test the obstacle avoidance performance and computational efficiency of the framework designed in this paper.
Then we generate a 6-DoF trajectory in a complex simulation environment through several narrow passages in succession 
and let the simulation model track it to verify the practicality of our framework.

We implement the trajectory optimization core algorithms in C++ using Eigen library and compile them using the C++17 standard.
The algorithm used to solve the optimization problem \eqref{equ:final_uncons_opt_problem} is L-BFGS \cite{liu1989limited}, 
with the BackTracking method used for line search.
The hardware platform used in the experiment is a Dell G5 laptop with Intel Core i7-10750H CPU@2.60GHz running Ubuntu 20.04 operating system.
The trajectory generation algorithms are all run serially on the CPU.

\begin{figure}[!t]
    \centering
    \subfloat[]{\includegraphics[width=3.5in]{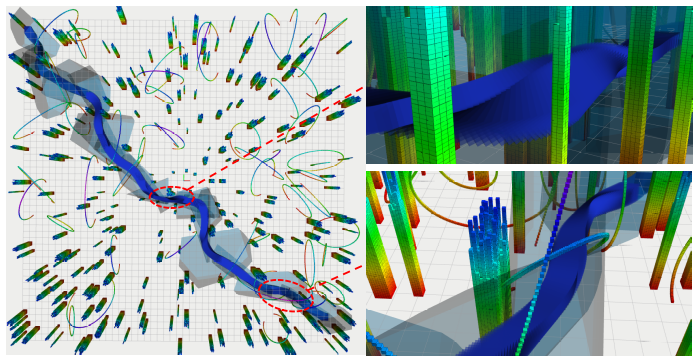}
    \label{fig:random_map_quat_planning.trajectory overview}}
    \hfil
    \subfloat[]{\includegraphics[width=3.5in]{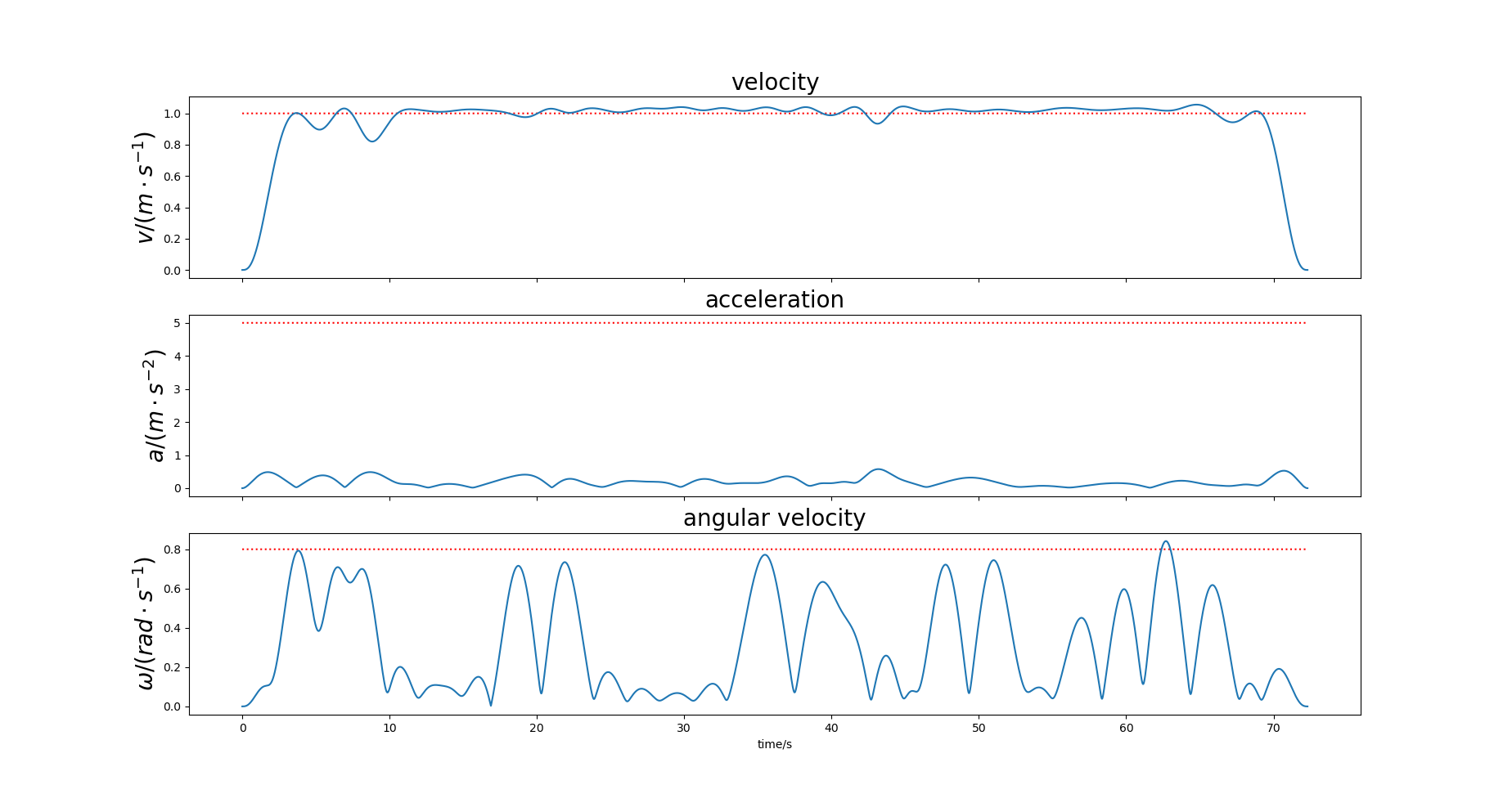}
    \label{fig:random_map_quat_planning.dyn}}
    \caption{6-DoF trajectory generated in a random map.(a) is the overview of the map and the trajectory, where a series of light blue transparent convex polyhedra form the safe flight corridor, 
    and the dark blue strip represents the volume swept by the cuboid of vehicle along the trajectory.(b) shows dynamical profiles of the trajectory, where The red dashed lines indicate the limited maximum values.}
    \label{fig:random_map_quat_planning}
\end{figure}

\begin{figure*}[!t]
    \centering
    \subfloat[]{\includegraphics[width=1.7in]{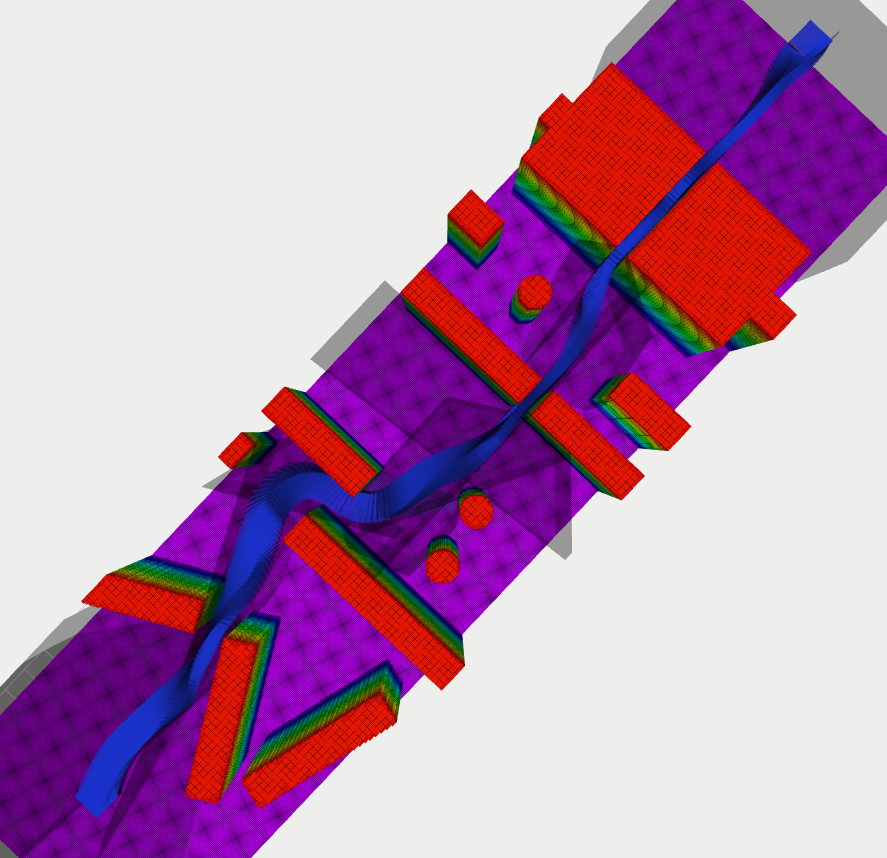}%
    \label{subfig:sfc_and_traj}}
    \hfil
    \subfloat[]{\includegraphics[width=1.7in]{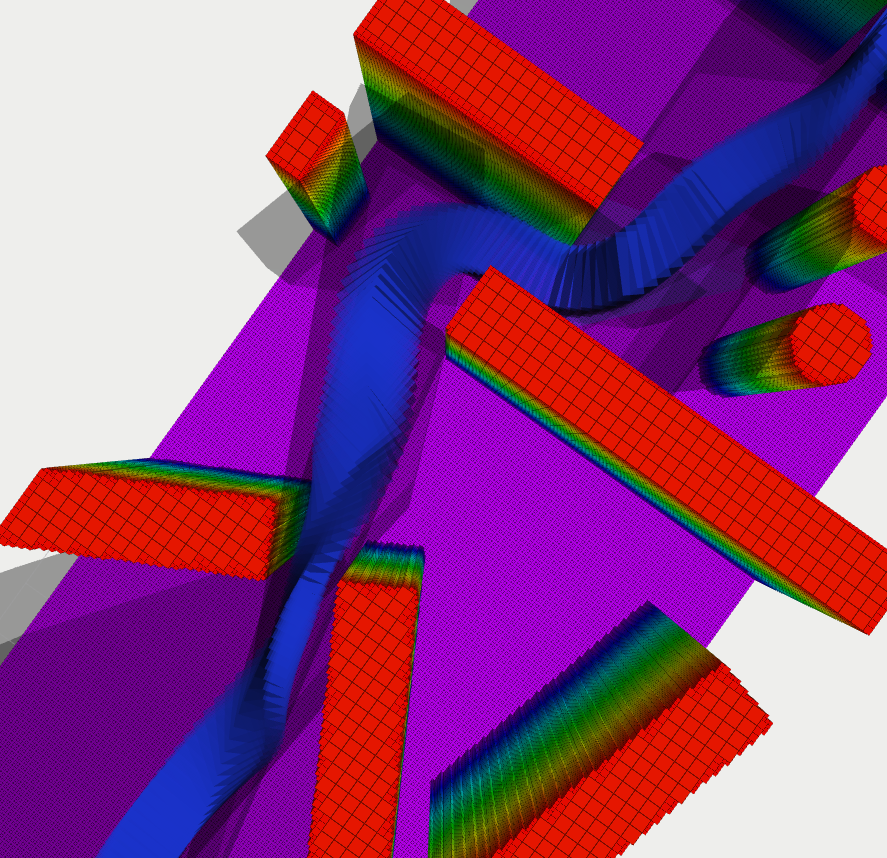}%
    \label{subfig:traj_detail_1}}
    \hfil
    \subfloat[]{\includegraphics[width=1.7in]{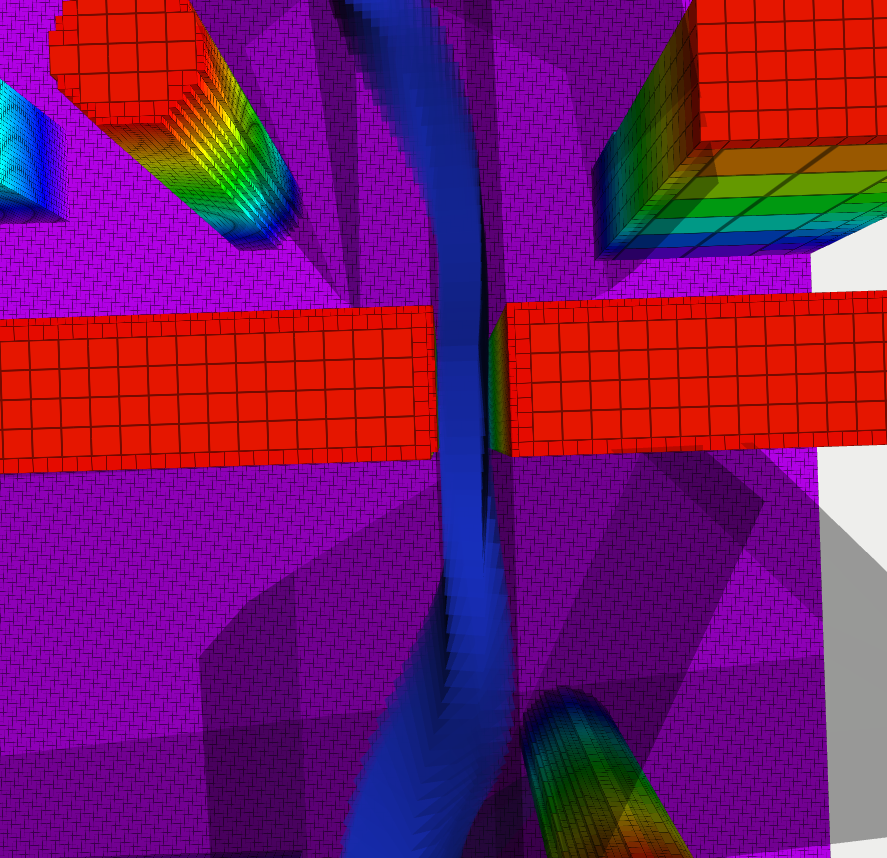}%
    \label{subfig:traj_detail_2}}
    \hfil
    \subfloat[]{\includegraphics[width=1.7in]{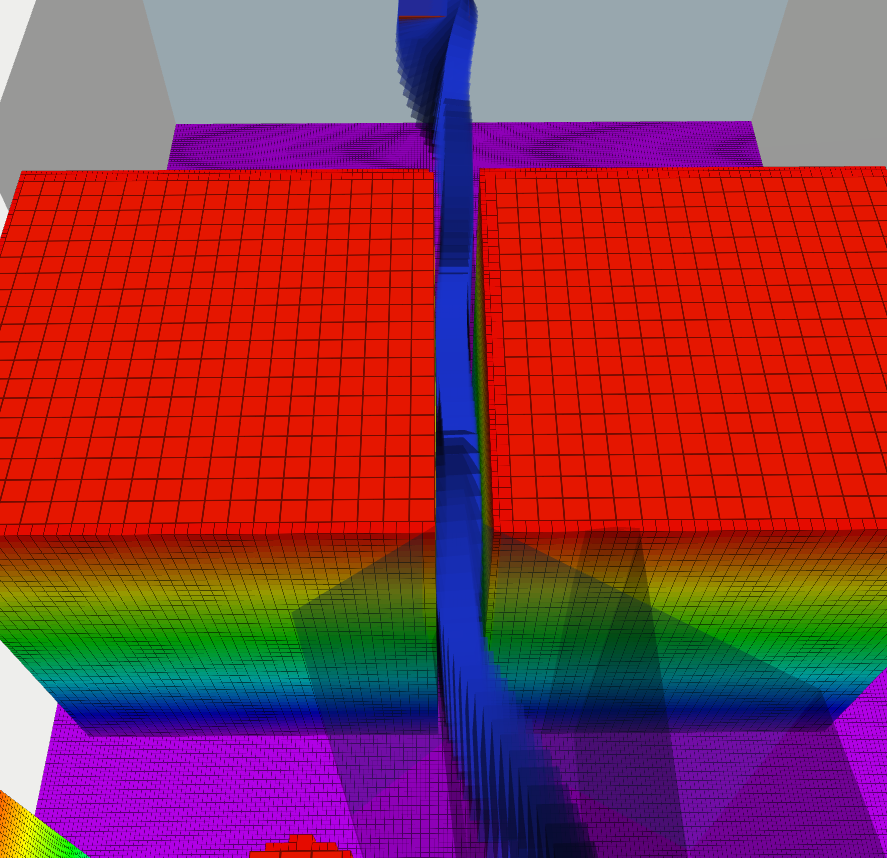}%
    \label{subfig:traj_detail_3}}
    \caption{The trajectory generated in a simulation environment with more dense obstacles.(a) is the overview of the generated trajectory;(b),(c), and (d) show details of the trajectory.}
    \label{fig:simulation_trajectory}
\end{figure*}

\begin{figure}[!t]
    \centering
    \includegraphics[width=3.5in]{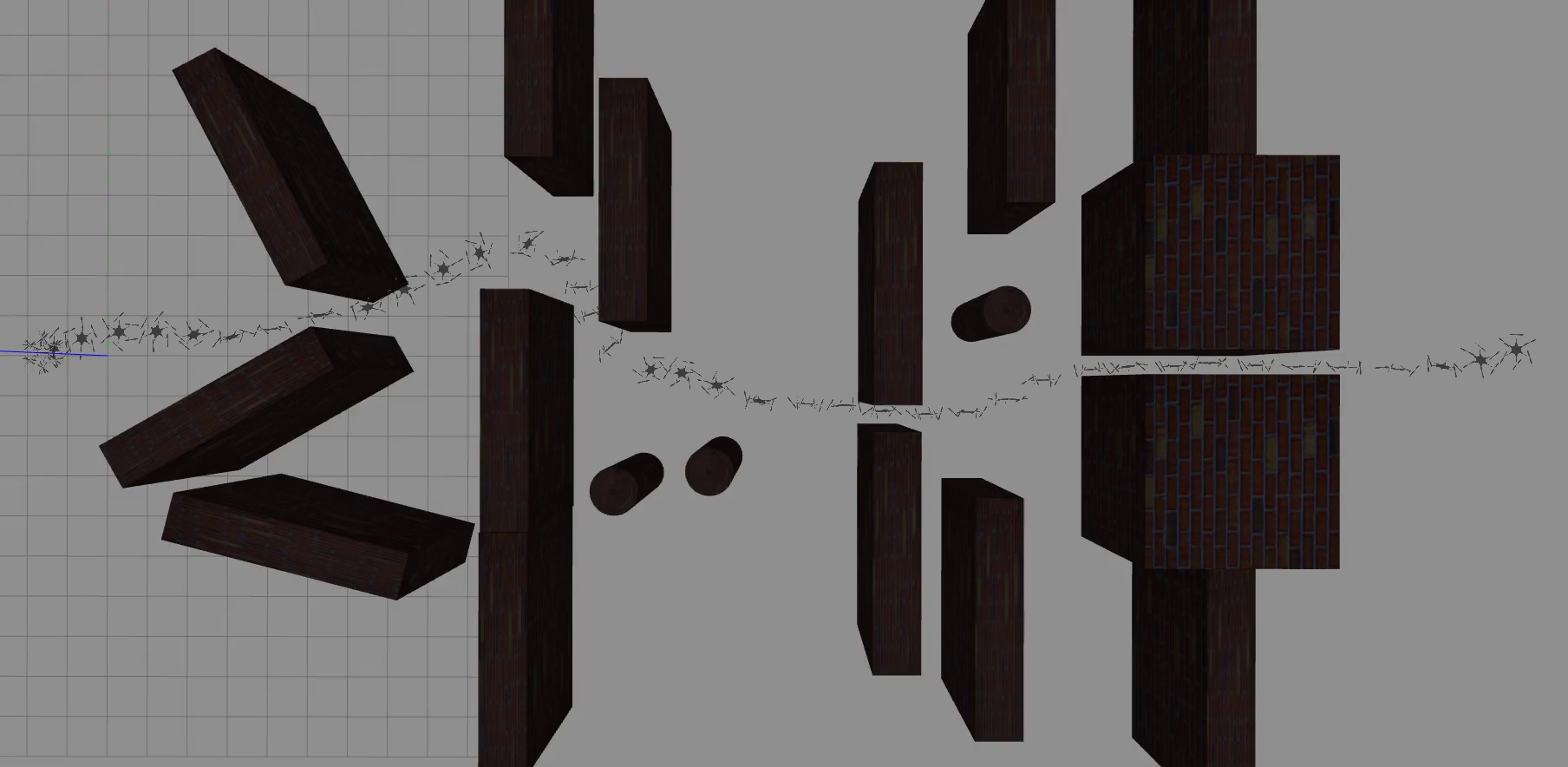}
    \caption{The actual flight trajectory.}
\label{fig:complex_env_simulation}
\end{figure}

\subsection{Preparation}
\begin{figure}[!t]
    \centering
    \includegraphics[width=3.5in]{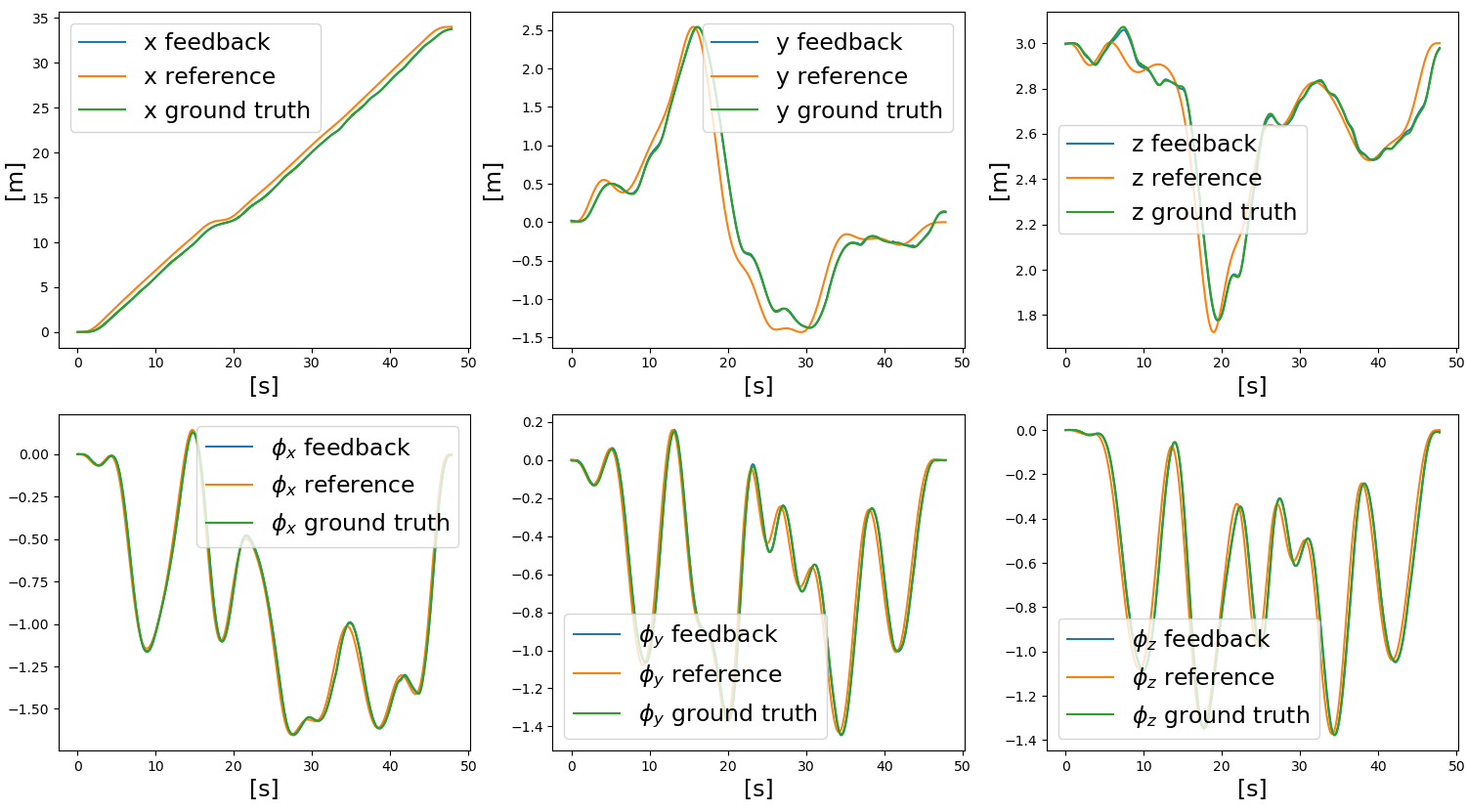}
    \caption{Tracking performance of the simulation model.The attitude is represented by Lie algebra $\phi$}
\label{fig:simulation_curves}
\end{figure}

The random maps generated in the trajectory generation experiment are $50\text{m}$ in width and length and $6\text{m}$ in height.
To achieve whole-body obstacle avoidance, as mentioned in \ref{subsec:safety_constraint}, we need to approximate the vehicle's shape as a convex polyhedron that wraps around the body.
Here we approximate it as a cuboid (Fig.~\ref{fig:approximate_cuboid}), 
whose dimensions are $l_x = l_y = 1.0\text{m},l_z = 0.35\text{m}$, 
The coordinates of its eight vertices in $\mathcal{F}_b$ are $\tilde{\bm{v}}_l = \begin{bmatrix}\pm \frac{l_x}{2} & \pm \frac{l_y}{2} & \pm \frac{l_z}{2}\end{bmatrix}^\text{T}$.
For parameter settings, we set the order of MINCO as $s=4$, that is, the order of the trajectory polynomial is $2s-1=7$;
The numerical integration resolution of penalty terms $\kappa=16$; 
The dynamics constraints are set to $v_\text{max}=0.8\text{m} \cdot \text{s}^{-1}, a_\text{max}=5.0\text{m} \cdot \text{s}^{-2}, \omega_\text{max}=0.8\text{rad} \cdot \text{s}^{-1}$;
The penalty weights are set to $\mathcal{W}_v=\mathcal{W}_a=\mathcal{W}_\omega=1\times10^4, \mathcal{W}_c=9\times10^4$.

\subsection{Trajectory Generation Results}
Fig.~\ref{fig:random_map_quat_planning} shows the 6-DOF trajectory planned and its kinematic properties under the above settings,
There are 300 cylindrical obstacles and 30 circular obstacles evenly distributed on the map.
As can be seen, the vehicle has been successfully constrained in the safe flight corridor throughout the whole journey and can flexibly change its attitude to avoid obstacles.
Fig.~\ref{fig:random_map_quat_planning.dyn} presents the constrained dynamical profiles.
It can be seen that although the imposed dynamic constraints are softened, they are effectively satisfied in the resulting trajectory.
Moreover, under the effect of the time regularization term, the trajectory speed reaches $v_{\text{max}}$ most of the time as shown in Fig.~\ref{fig:random_map_quat_planning.dyn}.

For computational efficiency, it takes an average of about 250ms to generate the trajectory connecting the diagonally opposite corners of the map as shown in Fig.~\ref{fig:random_map_quat_planning} (excluding the front-end path finding and SFC generation time).
The time complexity of each iteration of the optimization process (the value of the objective function and the related gradient should be calculated) is $O(M)$, so the average time of each segment $t_{\text{opt}}/M$ can be used to measure the computational efficiency.
We repeats trajectory generation hundreds of times in random maps with different obstacle distribution densities ($\kappa=16$), 
The average value of $t_{\text{opt}}/M$ is 10.7ms, which is not inferior to the state-of-the-art quadrotor $SE(3)$ planning framework \cite{han2021fast} on CPU.
Furthermore, as shown in \cite{han2021fast}, the computation of each subterm in penalty terms is independent of each other, which is also the case in this paper. 
Therefore, parallel computing can be used to significantly accelerate the computation efficiency of our framework, which can be achieved in many existing onboard computers with GPU.

\subsection{Simulation Results}
In this section, we design a simulation environment in which the trajectory needs to pass through a series of narrow areas to get from the start to the end, 
to test the performance of the designed framework in more extreme environments.
Then we let the simulation model of OmniHex track the generated trajectory to verify the feasibility of the designed framework.

Compared with the real OmniHex, the simulation model only keeps the key mechanical structures (e.g., the propellers and tilt-rotor units) to reduce the computational burden during simulation.
The simulator we use is Gazebo with PX4 Autopilot, and PID is used to control OmniHex.

Fig.~\ref{fig:simulation_trajectory} shows the generated trajectory.
It can be seen that the proposed algorithm could still obtain a collision-free and smooth trajectory even in such a complex environment, 
which shows the excellent adaptability of our method.
Also, as shown in Fig.~\ref{fig:complex_env_simulation} and Fig.~\ref{fig:simulation_curves}, the vehicle model tracks the trajectory accurately.
It avoids all obstacles and smoothly passes through narrow regions by flexibly changing the attitude, which is impossible for the traditional underactuated multirotor vehicle.
The simulation performance indicates that our method can be applied to the real world (as long as dynamic constraints and controller of the vehicle are appropriately designed so that the vehicle can accurately track the trajectory) and is capable of giving full play to the obstacle avoidance advantage of OMAVs.

\section{Conclusion}
In this paper, a 6-DOF trajectory generation framework is designed, which is computationally efficient, adaptable and can fully exploit the obstacle avoidance potential of omnidirectional multirotor vehicle.
A geometrically constrained whole body collision-free 6-DoF trajectory optimization problem is formulated;
A rational quaternion-based attitude parameterization method is adopted to obtain efficient optimization and high  quality soluton.
Simulation experiments are conducted to test the excellent performance of our framework.
We hope to build an omnidirectional multirotor vehicle with autonomous navigation and obstacle avoidance ability in complex environments by using this framework and the airborne sensing system, so as to promote the application of omnidirectional multirotor vehicle in aerial manipulators and complex environment search and rescue in disaster areas.
This will be our future work.



\begin{thebibliography}{99}
        \bibitem{brescianini2016design} D. Brescianini, and R. D’Andrea. "Design, modeling and con-trol of an omni-directional aerial vehicle,"\textit{in 2016 IEEE international conference on robotics and automation (ICRA)}. IEEE, 2016, pp. 3261–3266.
        \bibitem{kamel2018voliro} M. Kamel, et al. "The voliro omniorientational hexacopter: An agile and maneuverable tiltable-rotor aerial vehicle." \textit{IEEE Robotics \& Automation Magazine}, vol. 25, no. 4, pp. 34-44, 2018.
        \bibitem{brescianini2018computationally} D. Brescianini and R. D’Andrea. "Computationally eﬀicient trajectory generation for fully actuated multirotor vehicles," \textit{IEEE Transactions on Robotics}, vol. 34, no. 3, pp. 555–571, 2018.
        \bibitem{morbidi2018energy} F. Morbidi, D. Bicego, M. Ryll, and A. Franchi. "Energy-eﬀicient trajectory generation for a hexarotor with dual-tilting propellers," in \textit{2018 IEEE/RSJ International Conference on Intelligent Robots and Systems (IROS)}, IEEE, 2018, pp. 6226–6232.
        \bibitem{pantic2021mesh} M. Pantic, L. Ott, C. Cadena, R. Siegwart, and J. Nieto. "Mesh manifold based riemannian motion planning for omnidirectional micro aerial vehicles,". \textit{IEEE Robotics and Automation Letters}, vol.6, no. 3, pp. 4790–4797, 2021.
        \bibitem{nguyen2016time} H. Nguyen and Q. Pham. "Time-optimal path parameterization of rigid-body motions: Applications to spacecraft reorientation," \textit{Journal of Guidance, Control, and Dynamics}, vol. 39, no. 7. pp. 1667–1671, 2016.
        \bibitem{gao2017gradient} F. Gao, Y. Lin, and S. Shen. "Gradient-based online safe trajectory generation for quadrotor flight in complex environments," in \textit{2017 IEEE/RSJ international conference on intelligent robots and systems (IROS)}, IEEE, 2017,  pp. 3681–3688.
        \bibitem{zhou2020ego} X. Zhou, Z. Wang, H. Ye, C. Xu, and F. Gao. "Ego-planner: An esdf-free gradient-based local planner for quadrotors," \textit{IEEE Robotics and Automation Letters}, vol. 6 ,no. 2, pp. 478–485, 2020.
        \bibitem{deits2015computing} R. Deits and R. Tedrake. "Computing large convex regions of obstacle-free space through semidefinite programming," in \textit{I}, Springer, 2015, pages 109–124.
        \bibitem{gao2020teach} F. Gao, L. Wang, B. Zhou, X. Zhou, J. Pan, and S. Shen. "Teach-repeat-replan: A complete and robust system for aggressive flight in complex environments," \textit{IEEE Transactions on Robotics}, vol. 36, no. 5, pp. 1526–1545, 2020.
        \bibitem{gao2019flying} F. Gao, W. Wu, W. Gao, and S. Shen. "Flying on point clouds: Online trajectory generation and autonomous navigation for quadrotors in cluttered environments," \textit{Journal of Field Robotics}, vol. 36, no. 4, pp. 710–733, 2019.
        \bibitem{wang2022geometrically} Z. Wang, X. Zhou, C. Xu, and F. Gao. "Geometrically constrained trajectory optimization for multicopters," \textit{IEEE Transactions on Robotics}, pp. 1–10, 2022.
        \bibitem{han2021fast} Z. Han, Z. Wang, N. Pan, Y. Lin, Ch. Xu, and F. Gao. "Fast-racing: An open-source strong baseline for se(3) planning in autonomous drone racing," \textit{IEEE Robotics and Automation Letters}, vol. 6, no. 4, pp. 8631–8638, 2021.
        \bibitem{liu22017planning} S. Liu, M. Watterson, K. Mohta, K. Sun, Su. Bhattacharya, C. J. Taylor, and V. Kumar. "Planning dynamically feasible trajectories for quadrotors using safe flight corridors in 3-d complex environments," \textit{IEEE Robotics and Automation Letters},vol. 2, no. 3, pp. 1688–1695, 2017.
        \bibitem{fliess1995flatness} M. Fliess, J. Lévine, P. Martin, and P. Rouchon. "Flatness and defect of non-linear systems: introductory theory and examples," \textit{International journal of control}, vol. 61, no. 6, pp. 1327–1361, 1995.
        \bibitem{mellinger2011minimumsnap} D. Mellinger and V. Kumar. "Minimum snap trajectory generation and control for quadrotors," in \textit{2011 IEEE International Conference on Robotics and Automation}, IEEE, 2011, pp. 2520–2525.
        \bibitem{terzakis2014quaternion} G. Terzakis, P. Culverhouse, G. Bugmann, et al. "On quaternion based parametrization of orientation in computer vision and robotics," \textit{Journal of Engineering Science and Technology Review}, vol. 7, no. 1, pp. 82–93, 2014.
        \bibitem{liu1989limited} D. C. Liu and J. Nocedal, “On the limited memory bfgs method for large scale optimization,” \textit{Mathematical programming}, vol. 45, no. 1-3, pp. 503–528, 1989.
\end{thebibliography}
\end{document}